\documentclass[10pt,twocolumn,letterpaper]{article}

\usepackage{iccv}
\usepackage{times}
\usepackage{epsfig}
\usepackage{graphicx}
\usepackage{amsmath}
\usepackage{amssymb}

\usepackage[breaklinks=true,bookmarks=false]{hyperref}

\iccvfinalcopy 

\ificcvfinal\fi
\begin{document}

\title{Adaptive Binarization for Weakly Supervised Affordance Segmentation}

\author{Johann Sawatzky\\
University of Bonn\\
{\tt\small sawatzky@iai.uni-bonn.de}
\and
Juergen Gall\\
University of Bonn\\
{\tt\small gall@iai.uni-bonn.de}
}

\maketitle

\begin{abstract}
	The concept of affordance is important to understand the relevance of object parts for a certain functional interaction. Affordance types generalize across object categories and are not mutually exclusive. This makes the segmentation of affordance regions of objects in images a difficult task. In this work, we build on an iterative approach that learns a convolutional neural network for affordance segmentation from sparse keypoints. During this process, the predictions of the network need to be binarized. In this work, we propose an adaptive approach for binarization and estimate the parameters for initialization by approximated cross validation. We evaluate our approach on two affordance datasets where our approach outperforms the state-of-the-art for weakly supervised
affordance segmentation.
	\end{abstract}
	\section{Introduction}
	\label{sec:introduction}

	Affordances are properties of regions of scenes or objects which indicate their relevance for a certain functional interaction. Examples are \textit{holdable} for the external part of a mug or \textit{drivable} for a road.
	Localizing affordances is therefore an important task for autonomous systems that interact with the environment \cite{Koppula2013} as well as assistive systems that support visually impaired people~\cite{blind}. 
	Segmenting affordance regions, however, is a more difficult task than classical semantic image segmentation, which focuses on objects or categories that summarize regions of similar appearance like sky or grass. 
	
	Affordances are not only much more fine-grained than object categories, they represent a more abstract concept that generalizes across object categories. This requires that an affordance segmentation approach recognizes affordance for a previously unseen object class. For instance, it should generalize \textit{cuttable} from the blades of scissors or knives to the blade of a saw. Furthermore, affordance segmentation is a multi-label segmentation problem since affordance regions spatially overlap. This is in contrast to classical semantic image segmentation where the categories are mutually exclusive. This is in particular for weakly supervised learning, as it is addressed in this work, a big challenge. 
	
	Since acquiring pixelwise segmentation masks for training is very time consuming, methods for weakly supervised learning have been proposed that learn to segment object categories either from image labels \cite{weak_deeplab,SEC} or keypoint annotations~\cite{WTP}. 
	Our work builds on~\cite{cvpr} where an approach for affordance segmentation has been proposed that uses only keypoint annotations as weak supervision for training. The approach employs an iterative approach alternating between updating the parameters of a convolutional neural network and estimating the unknown segmentation masks of the training images. During this process, the predictions of the network need to be binarized. Since thresholding at the 50\% decision boundary, as it is done in a fully supervised setting, does not work for weakly supervised learning, an additional binary segmentation step is used in~\cite{cvpr}. 
	
	In this work, we propose an adaptive approach that determines the threshold for binarization for each training image and affordance class. Our approach not only avoids the additional segmentation step used in~\cite{cvpr} but also increases the affordance segmentation accuracy substantially. Since the initialization of the affordance segments based on the keypoints has a high impact on the accuracy, we show further how the parameters for initialization can be determined by cross validation using an approximation of the Jaccard index based on the given keypoints.         
	We evaluate our approach on the CAD 120 affordance dataset~\cite{cvpr} and the UMD part affordance dataset~\cite{UMD} using two different network architectures. In all settings, our approach outperforms~\cite{cvpr}. On the CAD 120 affordance dataset, the mean accuracy is increased by up to 17 percentage points compared to~\cite{cvpr}.

	\section{Related Work}
	\label{sec:related_work}
	Our work is related to affordance modeling as well as weakly supervised semantic segmentation methods. An affordance is an attribute of an object part that implies the possible usage of this object.
	Assigning affordances to object parts is not trivial, while \cite{cvpr} and \cite{UMD} simply let a human annotator decide, others use more sophisticated statistics like mining of word co-occurrences \cite{Chao2015} or object attribute graph structures \cite{Zhu2014}.\\ 
	Modeling affordances can be the final goal or an intermediate step. In \cite{Castellini2011}, affordances of an object are defined in terms of hand poses during interaction. These affordances are used along with object appearance features for object classification. \cite{Kjellstroem2011} apply implicit affordance modeling for simultaneous hand action and object detection. While \cite{Zhu2015} combine object affordances with physical observables and human pose to obtain a generative model for object tasks, \cite{Koppula2016} use object affordance, object appearance and human poses for action detection.\\
	Since recognizing affordances is crucial for the constructive manipulation of objects by robots, several approaches that require full supervision have been proposed.  
	While some use geometric information like orientation of object surfaces \cite{Katz2014}, 3d point clouds \cite{Kim2014}, or normal and curvature features \cite{UMD}, others rely only on appearance. 
	\cite{Hermans2011} predict attributes from appearance and affordances from attributes, \cite{Desai2013} measure similarity between query and training image by the location of object parts, \cite{cvpr} train a deep model on RGB data. RGB-D data is exploited by \cite{Lenz2015} who propose a two stage cascade to model graspable regions and \cite{Roy2016} who train a CNN to predict depth information and affordances from RGB data simultaneously.
	CNNs were also used in \cite{Nguyen2016} for a pixelwise affordance segmentation in RGB-D data and in \cite{Kumra2017} to predict grasps. \cite{Grabner2011,Jiang2013,Koppula2014} exploit human poses to localize object affordances.
	
	Weakly supervised semantic image segmentation faced rapid progress in recent time. 
	\cite{weak_deeplab} use an expectation-maximization (EM) approach with area constraints to train a CNN. \cite{WTP} use keypoint annotations and incorporate objectness into their loss function, \cite{SEC} exploit localization cues from an image level classifier, area constrains and CRFs. \cite{Hou2016} rely on superpixels and \cite{saliency_segmentation} learn a model from image labels and saliency predictions. In \cite{Durand2017}, an approach based on pooling of classwise heat maps along with image labels was proposed. A weakly supervised affordance segmentation approach based on EM similar to \cite{weak_deeplab} was proposed in \cite{cvpr}. Our approach builds on this work.
	
	\section{Weakly Supervised Affordance Segmentation}
	\label{sec:Methodology}
Our approach for weakly supervised affordance segmentation extends the approach~\cite{cvpr} by adaptive binarization and approximated cross validation for estimating hyperparameters. We therefore briefly describe~\cite{cvpr} first and then describe in Section~\ref{subsec:adaptive_thresholding} the adaptive binarization and in Section~\ref{subsec:approximated_cross_validation} approximated cross validation.  

	\subsection{Method}
	\label{sec:Methodology_a}
The approach~\cite{cvpr} extends fully convolutional neural networks like \cite{deeplabVGG} or \cite{deeplabResnet} for the task of affordance segmentation. In contrast to semantic image segmentation, where only one label per pixel needs to be predicted, affordance segmentation requires to predict a set of labels per pixel since an object region might contain multiple affordance types. The approach predicts $P(Y|I;\theta)$ where $I$ denotes the input image, $\theta$ denotes the parameters of the model, i.e.\ the weights of the neural network, and $Y = \{y_{i,l}\}$ with $y_{i,l} \in \mathcal \{0,1\}$ is the pixelwise segmentation. If $y_{i,l}=1$ the affordance type $l$ is predicted for pixel $i$. 
Due to the multi-label problem, the network uses a sigmoid layer instead of a softmax layer~\cite{deeplabVGG,deeplabResnet}:
	\begin{align} \label{eq:supervised_dcnn_pixel_label}
	P(y_{i,l}=1|I;\theta) = \frac{1}{1+\exp\left(-g_{i,l}\left(y_{i,l}|I;\theta\right)\right)},
	\end{align}
where $g_{i,l}$ is the value of the previous layer of the neural network.
For segmentation, the predicted probabilities $P(y_{i,l}|I;\theta)$ need to be binarized. In \cite{cvpr}, this is achieved by the standard 50\% threshold:   
	\begin{align}\label{eq:thres}
	\hat{y}_{i,l} = \begin{cases} 1 & \text{if } P(y_{i,l}=1|I;\theta) \geq 0.5 \\0&\text{otherwise.}\\ \end{cases}
	\end{align}

The model parameters $\theta$ are determined during training. In the strongly supervised setting, training means optimizing the log-likelihood: 
	\begin{align}\label{eq:J}
	\centering
	J(\theta) = \log P(Y|I;\theta) = \sum_{i=1}^{n} \sum_{l\in\mathcal{L}} \log P(y_{i,l}|I;\theta).
	\end{align}
In the weakly supervised setting, the log-likelihood can not be calculated since $Y$ is not given during training. In~\cite{cvpr}, it was proposed to train the model only from a set of keypoints $Z=\{(l_k,i_k)\}$, which denote the presence of the affordance $l_k$ at pixel $i_k$, using expectation-maximization (EM). During training, both $Y$ and $\theta$ need to be estimated from $Z$. The approach starts with an initial estimate $\hat{Y}$, which is derived from the keypoints $Z$ by labeling all pixels within a radius of $\sigma$ around a keypoint:    
\begin{align}\label{eq:sigma}
\hat{y}_{i,l} = \begin{cases} 1 & \text{if } \vert \{ (l_k,i_k){\in}Z : l_k{=}l \land \vert i_k-i\vert{\leq}\sigma \}\vert > 0 \\0&\text{otherwise.}\\ \end{cases}
\end{align}
In contrast to \cite{cvpr} that uses fixed values for initialization, we discuss in Section \ref{subsec:approximated_cross_validation} how $\sigma$ can be estimated by approximated cross validation. 

\begin{figure*}
	{
\centering
\includegraphics[height=90mm]{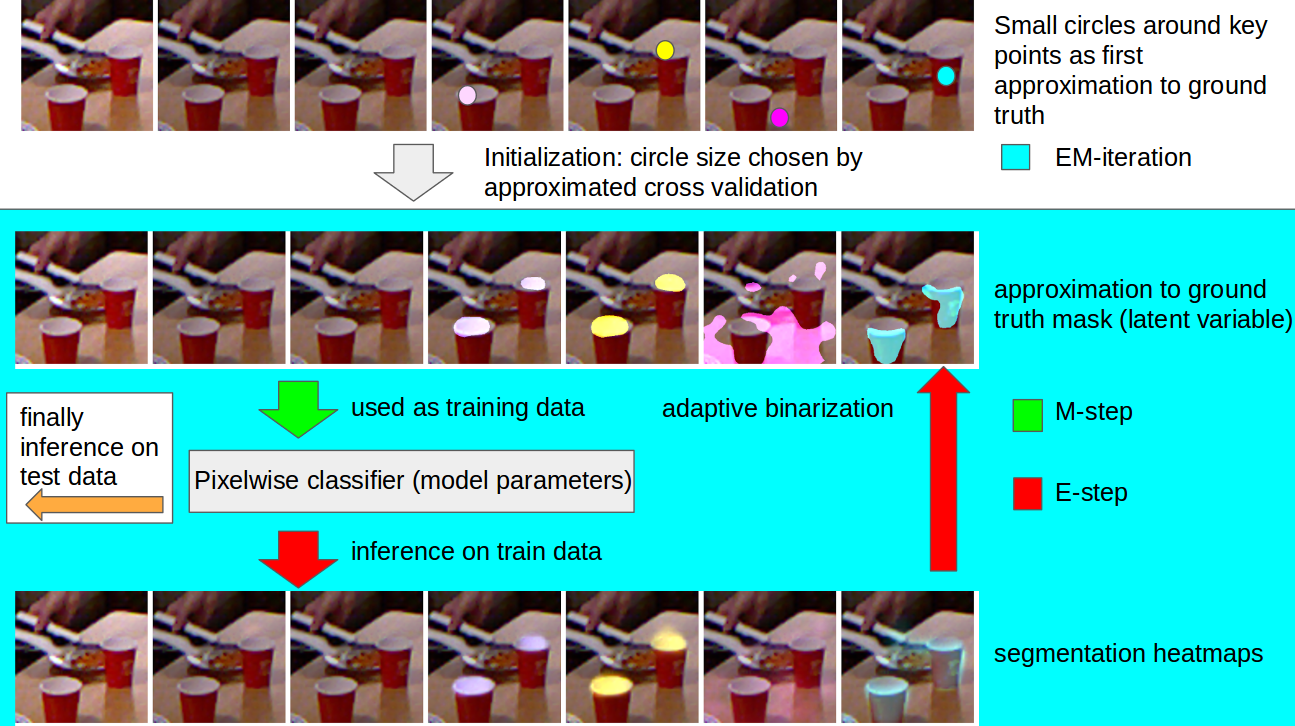}
\caption{\label{fig:pipeline} 
Illustration of our approach for affordance segmentation using key points as weak supervision. The CNN is trained by iteratively updating the segmentation masks for the training images (E-step) and the parameters of the network (M-step).}       
}
\end{figure*}

After $\hat{Y}$ is estimated, the weights of the network $\theta$ are updated by optimizing $J(\theta) = \log P(\hat{Y}|I;\theta)$. Given the new weights $\theta$, the CNN predicts $P(y_{i,l}|I;\theta)$ for each training image and $\hat{Y}$ is refined by binarization of the CNN predictions. The 50\% threshold used in~\eqref{eq:thres}, however, is only valid for the fully supervised setting. While in \cite{cvpr} an additional GrabCut step is used to address this issue, we propose an adaptive approach that determines the threshold for binarization for each training image and affordance class. This not only increases the accuracy, but it also reduces the training time since an additional GrabCut step is not needed anymore by our approach. The approach for adaptive binarization is discussed in Section~\ref{subsec:adaptive_thresholding}. Our weakly supervised approach for affordance segmentation is illustrated in Figure~\ref{fig:pipeline}.

To reduce overfitting and perform approximated cross validation as described in Section~\ref{subsec:approximated_cross_validation}, we split the training set into three equally sized subsets A, B, and C. During the M-step, we train the convolutional network on each of the tuples (A,B), (B,C), and (C,A). During the E-step, each network predicts $P(y_{i,l}|I;\theta)$ for the set that was not used for training. As in \cite{cvpr}, we use two EM iterations to obtain $\hat{Y}$ for all training images. The final CNN model is then obtained by optimizing $J(\theta) = \log P(\hat{Y}|I;\theta)$ on the entire training set.

\subsection{Adaptive Binarization}
\label{subsec:adaptive_thresholding}
We first want to explain why the binarization as described in \ref{eq:thres} is not optimal for the weakly supervised case. Let us first consider an optimal classifier that separates two classes perfectly in the training data. In this case, $P(y_{i,l}=1|I;\theta)\geq 0.5$ if a pixel is annotated by $y_{i,l}=1$ and $P(y_{i,l}=1|I;\theta)< 0.5$ if it is annotated by $y_{i,l}=0$. 
Hence, using 50\% as threshold for binarization is optimal. For weakly supervised learning, $\hat{Y}$ is in particular after the initialization only a poor estimation of the unknown ground truth segmentation masks $Y$ of the training data such that $\hat{y}_{i,l}\neq y_{i,l}$ for many pixels. This means that the optimal threshold is unknown. However, we can use the keypoints $Z$ to obtain an estimate of the threshold:       				
        \begin{align}
        \label{eq:threshold}
	    {\hat{y}}_{i,l} = \begin{cases} 1 & \text{if } P(y_{i,l}=1|I;\theta) \geq t \\0&\text{otherwise}\\ \end{cases}
	\end{align}
	where
	\begin{align}
        \label{eq:threshold2}
	    t = \min\left\{0.5,f\left(\{P(y_{i_k,l}=1|I;\theta)\}_{l_k=l}\right)\right\}. 
	\end{align}
$P(y_{i_k,l}=1|I;\theta)\}_{l_k=l}$ are the predictions of the classifier for all keypoints in the training image $I$ with label $l$ and $f$ computes either the mean or median of \mbox{$P(y_{i_k,l}=1|I;\theta)\}_{l_k=l}$}. In our default experimental setting, we will have only one keypoint for each affordance occurring in an image. In general, one can expect that the threshold is below $0.5$ since 
the ratio $\tfrac{\vert\{i:y_{i,l}=0 \land \hat{y}_{i,l}=1\}\vert}{\vert\{i:y_{i,l}=0\}\vert}$ is usually lower than $\tfrac{\vert\{i:y_{i,l}=1 \land \hat{y}_{i,l}=0\}\vert}{\vert\{i:y_{i,l}=1\}\vert}$. As soon as the threshold reaches $0.5$, we can replace the adaptive threshold by $0.5$. We therefore limit the threshold by $0.5$.

	\subsection{Approximated Cross Validation}
	\label{subsec:approximated_cross_validation}
		In the fully supervised setup, hyperparameters can be optimized by cross-validation on the training set using the same measure that is also used for evaluation.  Since the ground truth masks $Y$, however, are unknown in the weakly supervised setup, exact cross validation is not possible. We therefore propose to approximate the Jaccard index, which measures the intersection over union between the ground-truth $Y$ and the prediction $\hat{Y}$, on the validation set. Since the Jaccard index is computed per affordance class $l$ and then averaged over all classes, we discuss only the binary case with $y_i \in \{0,1\}$. Let $P(y_{i}=1) = \tfrac{\vert\{i:y_{i}=1\}\vert}{\vert\{i\}\vert}$ be the unknown percentage of pixels with $y_i = 1$ and $P(\hat{y}_{i}=1) = \tfrac{\vert\{i:\hat{y}_{i}=1\}\vert}{\vert\{i\}\vert}$ the known percentage of pixels that have been classified with $\hat{y}_i = 1$. 
We can approximate $P(\hat{y}_{i}=1|y_{i}=1)$ by measuring how often a keypoint annotated by the affordance class has been correctly classified. Similarly, $P(\hat{y}_{i}=1|y_{i}=0)$ is given by the percentage of keypoints that have been misclassified. This gives the relation   		
\begin{align}
P(\hat{y}_{i}=1) =& P(\hat{y}_{i}=1|y_{i}=1)P(y_{i}=1) \\ \nonumber
                  &+ P(\hat{y}_{i}=1|y_{i}=0)(1-P(y_{i}=1))
\end{align}
and thus
\begin{align}
P(y_{i}=1)=\frac{P(\hat{y}_{i}=1)-P(\hat{y}_{i}=1|y_{i}=0)}{P(\hat{y}_{i}=1|y_{i}=1)-P(\hat{y}_{i}=1|y_{i}=0)}.
\end{align}
The Jaccard index which is  
\begin{align}
J=\frac{ \vert\{i:y_{i}=1 \land \hat{y}_{i}=1\}\vert}{\vert\{i:y_{i}=1\}\vert + \vert\{i:y_{i}=0 \land \hat{y}_{i}=1\}\vert}
\end{align}
can then be approximated by      
\begin{align}\label{eq:approxJ}
J_{approx} = \frac{P(\hat{y}_{i}=1|y_{i}=1)P(y_{i}=1)}{P(y_{i}=1)+P(\hat{y}_{i}=1|y_{i}=0)(1-P(y_{i}=1))}.
\end{align}
As mentioned in Section~\ref{sec:Methodology_a}, we split the training set into three subsets for approximate cross-validation.

\begin{table*}
\centering
\begin{tabular}{lllllllll}
\hline
CAD 120 & Bck & Open & Cut & Contain & Pour & Support & Hold & Mean \\ \hline
non-adaptive (VGG) & 0.62 & 0.09 & 0.20 & 0.41 & 0.35 & 0.11 & 0.40 & 0.31  \\ \hline
adaptive (VGG) & 0.68 & 0.10 & 0.23 & 0.44 & 0.36 & 0.50 & 0.47 & 0.40  \\ \hline
\end{tabular}
\caption{\label{tab:cad_nat} Comparison of adaptive binarization with non-adaptive binarization. The Jaccard index is reported for the object split of CAD 120 affordance dataset. }
\end{table*}

\begin{table*}
\centering
\begin{tabular}{lllllllll}
\hline
UMD & Grasp & Cut & Scoop & Contain & Pound & Support & Wgrasp & mean\\ \hline
non-adaptive (VGG) & 0.32 & 0.12 & 0.48 & 0.46 & 0.08 & 0.33 & 0.69 & 0.36\\ \hline
adaptive (VGG) & 0.31 & 0.18 & 0.56 & 0.49 & 0.08 & 0.41 & 0.66 & 0.38\\ \hline
\end{tabular}
\caption{\label{tab:umd_nat} Comparison of adaptive binarization with non-adaptive binarization. The Jaccard index is reported for the novel split of the UMD part affordance dataset.}
\end{table*}

	\section{Experiments}
	\label{sec:Experiments}
	
	For evaluation, we use the CAD 120 affordance dataset~\cite{cvpr} and the UMD part affordance dataset~\cite{UMD}. We use the splits separating the object classes (novel split on UMD and object split on CAD) and the splits which do not separate the object classes (category split on UMD and actor split on CAD). As measure, we use the Jaccard index. We report the results using the VGG architecture \cite{deeplabVGG} and the ResNet architecture \cite{deeplabResnet} as underlying convolutional network.
	First, we conduct ablation experiments to show the impact of our two key components, adaptive binarization and approximated cross validation. Second, we compare our approach with other weakly supervised segmentation approaches. If not otherwise specified, we use our approach based on the VGG architecture with adaptive binarization and approximate cross validation to determine $\sigma$~\eqref{eq:sigma}. As in~\cite{cvpr}, we use one keypoint per affordance class and training image. In Section~\ref{sec:keypoints}, we also evaluate the impact of the number of keypoints.


\begin{table*}
\centering
\begin{tabular}{lllllllll}
\hline
CAD 120 & Bck & Open & Cut & Contain & Pour & Support & Hold & Mean \\ \hline
Max thres. 1.0 (VGG) & 0.62 & 0.08 & 0.21 & 0.34 & 0.33 & 0.39 & 0.19 & 0.31  \\ \hline
Max thres. 0.5  (VGG) & 0.68 & 0.10 & 0.23 & 0.44 & 0.36 & 0.50 & 0.47 & 0.40  \\ \hline
\end{tabular}
\caption{\label{tab:cad_min} 
Impact of limiting the adaptive threshold \eqref{eq:threshold} by 0.5. The Jaccard index is reported for the object split of the CAD 120 affordance dataset.
}
\end{table*}

\begin{table*}
\centering
\begin{tabular}{lllllllll}
\hline
UMD & Grasp & Cut & Scoop & Contain & Pound & Support & Wgrasp & mean\\ \hline
Max thres. 1.0 (VGG) & 0.32 & 0.04 & 0.36 & 0.42 & 0.05 & 0.23 & 0.64 & 0.29\\ \hline
Max thres. 0.5 (VGG) & 0.31 & 0.18 & 0.56 & 0.49 & 0.08 & 0.41 & 0.66 & 0.38\\ \hline
\end{tabular}
\caption{\label{tab:umd_min} 
Impact of limiting the adaptive threshold \eqref{eq:threshold} by 0.5. The Jaccard index is reported for the novel split of the UMD part affordance dataset.}
\end{table*}

\subsection{Adaptive Binarization}
\label{subsec:minimum}
First we evaluate the impact of adapting the binarization to each training image and affordance class in comparison to using a constant threshold for each affordance class. To this end, instead of using $\min\left\{0.5,f\left(\{P(y_{i_k,l}=1|I;\theta)\}_{l_k=l}\right)\right\}$ as an individual threshold for each image $I$, we take the average of these thresholds over all images in the training set labeled with the affordance class $l$. Note that $f\left(\{P(y_{i_k,l}=1|I;\theta)\}_{l_k=l}\right){=}P(y_{i_k,l}=1|I;\theta)$ in this experiment since we use only one keypoint per affordance $l$ and image $I$.  

The results for the object split of the CAD 120 affordance dataset and the novel split of the UMD part affordance dataset are shown in Tables \ref{tab:cad_nat} and \ref{tab:umd_nat}, respectively. Compared to the proposed adaptive binarization approach, the accuracy decreases for all affordance classes and the background, which are regions annotated without any affordance class. In average, the accuracy decreases by $-9\%$. On UMD, the decrease is smaller but still $-2\%$. The effect on CAD is larger since the size of the affordance regions varies more across the training images in comparison to UMD.


As discussed in Section~\ref{subsec:adaptive_thresholding}, we limit the adaptive threshold by $0.5$, which is the optimal threshold for a fully supervised trained model. Tables \ref{tab:cad_min} and \ref{tab:umd_min} show the results when the threshold is not limited, i.e., the adaptive threshold can even get close to one. As expected, the accuracy drops for both datasets by $-9\%$ since a threshold above $0.5$ would produce even in the fully supervised case too small affordance segments.

%
%
%

\subsection{Approximated Cross Validation}
The initialization depends on the value $\sigma$, which determines the initial affordance segments around the keypoints~\eqref{eq:sigma}. This is shown in the last column of Table \ref{tab:approximated_cross_validation_CAD} where we report the mean Jaccard index for three values of $\sigma$. Note that $\sigma$ is set proportional to the image width $w$. The results show that the accuracy strongly depends on the initialization. The strongest variation can be observed for the category split of the UMD part affordance dataset where the accuracy varies between $0.44$ to $0.61$. The approximated Jaccard index computed from the keypoints in the training set, which is reported in the second column of Table \ref{tab:approximated_cross_validation_CAD}, however, correlates with the Jaccard index on the test set. This shows that using approximated cross validation to determine $\sigma$ works very well in practice. Note that the values between the Jaccard index and its approximation differ since the first measure is computed over the test set and the second over the training set. In all experiments except of Table \ref{tab:approximated_cross_validation_CAD}, we have determined $\sigma$ by approximated cross validation.

	
	\begin{table*}
	\centering
	\begin{tabular}{l|lll||lll}
	& \multicolumn{3}{c||}{approx.\ Jaccard train} & \multicolumn{3}{c}{Jaccard test} \\ \hline
	$\sigma$ relative to image width\; & 0.03$w$ & 0.06$w$ & 0.12$w$ & 0.03$w$ & 0.06$w$ & 0.12$w$  \\ \hline \hline
	CAD actor split & 0.38 & \textbf{0.40} & 0.30 & 0.41 & \textbf{0.42} & 0.37   \\ \hline	
	CAD object split  & 0.48 & \textbf{0.50} & 0.39 & 0.38 & \textbf{0.40} & 0.35   \\ \hline
	UMD category split & 0.57 & \textbf{0.58} & 0.44 & \textbf{0.61} & 0.59 & 0.44   \\ \hline
	UMD novel split & \textbf{0.66} & 0.62 & 0.44 &  \textbf{0.38} & \textbf{0.38} & 0.35   \\ \hline
	 \end{tabular}
	 \caption{\label{tab:approximated_cross_validation_CAD} Impact of $\sigma$ \eqref{eq:sigma}. The second column contains the approximated Jaccard index \eqref{eq:approxJ} computed on the training data for three values of $\sigma$. The approximated Jaccard index is used to determine $\sigma$. The third column contains the Jaccard index computed on the test data for three values of $\sigma$.} 
	\end{table*}

	
	\begin{figure*}[t]
	{
\centering
\includegraphics[height=53mm]{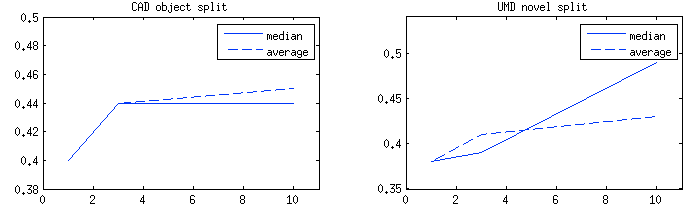}
\caption{\label{fig:multiple_points} 
Affordance segmentation with more than one keypoint per image and affordance. For the function $f$ \eqref{eq:threshold}, we compare average and median. The mean Jaccard index is plotted over the number of keypoints.}        
}
\end{figure*}

\subsection{Varying Number of Keypoints}\label{sec:keypoints}
Our approach also works with multiple keypoints per affordance class in an image. In this case, we compare two functions for $f\left(\{P(y_{i_k,l}=1|I;\theta)\}_{l_k=l}\right)$  \eqref{eq:threshold}, namely taking the average or the median of $\{P(y_{i_k,l}=1|I;\theta)\}_{l_k=l}$.    
The results are reported in Figure \ref{fig:multiple_points}. For the object split of the CAD 120 affordance dataset, average and median perform similar and the accuracy increases only slightly after three keypoints. A similar behavior can be observed for the novel split of the UMD part affordance dataset, but the accuracy increases more if the median is used.            


\subsection{Comparison to the State-of-the-art}
\begin{figure*}
	{
\centering
\includegraphics[height=90mm]{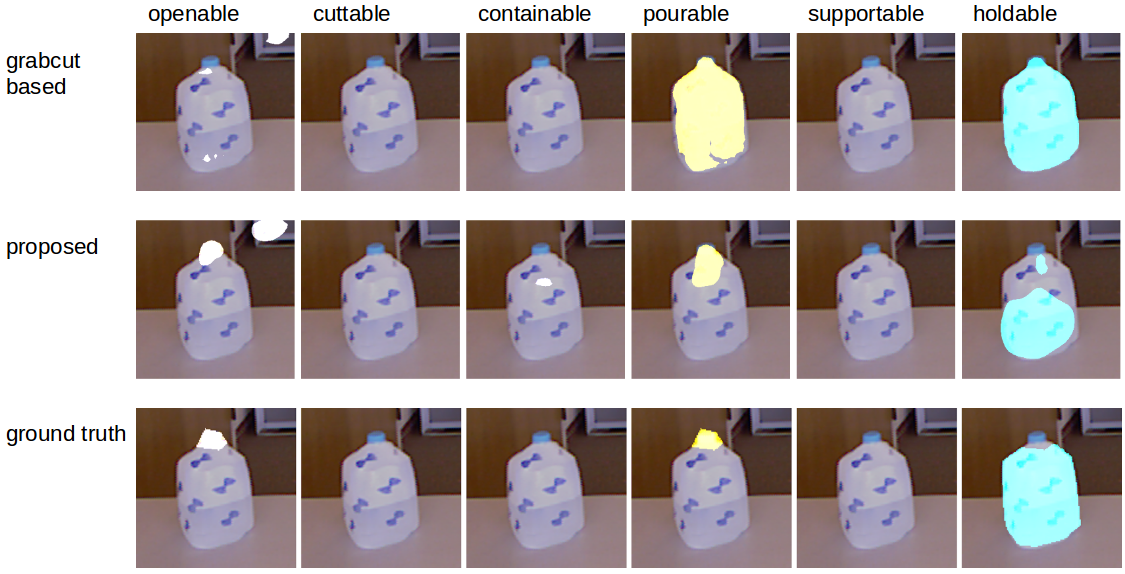}
\caption{\label{fig:example} 
Qualitative comparison of our approach (second row) with \cite{cvpr} (first row). Our approach localizes even small affordance parts while the GrabCut step in \cite{cvpr} merges the cap with the entire object.}       
}
\end{figure*}

We finally compare our approach with other weakly supervised semantic segmentation approaches~\cite{SEC,WTP,weak_deeplab,cvpr}. The results for both splits on the CAD 120 affordance dataset are reported in Table~\ref{tab:cad120_dataset_results}, while the results for the UMD part affordance dataset are reported in Table \ref{tab:umd_dataset_results}. The methods~\cite{SEC,weak_deeplab} use only image labels and therefore weaker supervision. It is therefore expected that methods that use more supervision in form of keypoints achieve a higher accuracy. For the methods \cite{WTP,cvpr} and our approach, we use one keypoint for each affordance class in an image. The parameter $\sigma$ has been determined by approximated cross validation. We also report the results as in \cite{cvpr} for the VGG architecture and the ResNet architecture. Our approach outperforms \cite{cvpr} and the other methods on both datasets. While our approach  achieves with the ResNet architecture on all datasets and splits a better mean accuracy than VGG, this is not the case for \cite{cvpr} where VGG is sometimes better. For the actor split of the CAD 120 affordance dataset, the mean accuracy is improved by $+17\%$ compared to \cite{cvpr}. This shows the benefit of adaptive binarization for weakly supervised affordance segmentation. 
Qualitative results are shown in Figure~\ref{fig:example}.


%

%

	\begin{table*}
	 \centering
	 \begin{tabular}{lllllllll}
	  CAD 120 & Bck & Open & Cut & Contain & Pour & Support & Hold & Mean \\ \hline\hline
 \multicolumn{9}{l}{image label supervision - actor split} \\ \hline\hline
Area constraints \cite{weak_deeplab} & 0.53 & 0.11 & 0.02 & 0.09 & 0.09 & 0.07 & 0.15 & 0.15 \\ \hline
SEC \cite{SEC} & 0.53 & 0.43 & 0.00 & 0.25 & 0.09 & 0.02 & 0.20 & 0.22 \\ \hline\hline
\multicolumn{9}{l}{keypoint supervision - actor split} \\ \hline\hline
WTP \cite{WTP} & 0.53 & 0.13 & 0.00 & 0.10 & 0.08 & 0.11 & 0.22 & 0.17 \\ \hline
 \cite{cvpr} (VGG) & 0.61 & 0.33 & 0.0 & 0.35 & 0.30 & 0.22 & 0.43 & 0.32  \\ \hline
Proposed (VGG) & 0.71 & 0.47 & 0.0 & 0.36 & 0.37 & 0.56 & 0.49 & 0.42  \\ \hline
\cite{cvpr} (ResNet) & 0.60 & 0.25 & 0.00 & 0.35 & 0.30 & 0.17 & 0.42 & 0.30  \\ \hline
 Proposed (ResNet) & 0.77 & 0.50 & 0.00 & 0.43 & 0.39 & 0.64 & 0.56 & 0.47  \\ \hline\hline
\multicolumn{9}{l}{image label supervision - object split} \\ \hline\hline
Area constraints \cite{weak_deeplab} & 0.59 & 0.03 & 0.03 & 0.01 & 0.02 & 0.02 & 0.28 & 0.14 \\ \hline
 SEC \cite{SEC} & 0.54 & 0.04 & 0.09 & 0.13 & 0.09 & 0.08 & 0.13 & 0.16 \\ \hline\hline
\multicolumn{9}{l}{keypoint supervision - object split} \\ \hline\hline
 WTP \cite{WTP} & 0.57 & 0.01 & 0.00 & 0.02 & 0.09 & 0.03 & 0.19 & 0.13 \\ \hline
 \cite{cvpr} (VGG) & 0.62 & 0.08 & 0.08 & 0.24 & 0.22 & 0.20 & 0.46 & 0.27  \\ \hline
 Proposed (VGG) & 0.68 & 0.10 & 0.23 & 0.44 & 0.36 & 0.50 & 0.47 & 0.40  \\ \hline
 \cite{cvpr} (ResNet) & 0.69 & 0.11 & 0.09 & 0.28 & 0.21 & 0.36 & 0.56 & 0.33  \\ \hline
  Proposed (ResNet) & 0.74 & 0.15 & 0.21 & 0.45 & 0.37 & 0.61 & 0.54 & 0.44  \\ \hline
	 \end{tabular}
	 \caption{\label{tab:cad120_dataset_results} Comparison of our method to the state-of-the-art on the CAD 120 affordance dataset. The Jaccard index is reported.}
	 \end{table*}
	 
	\begin{table*}
	 \centering
	 \begin{tabular}{lllllllll}
	 UMD & Grasp & Cut & Scoop & Contain & Pound & Support & Wgrasp & mean\\ \hline\hline
	 \multicolumn{9}{l}{image label supervision - category split} \\ \hline\hline
	  Area constraints \cite{weak_deeplab} & 0.06 & 0.04 & 0.10 & 0.14 & 0.22 & 0.04 & 0.37 & 0.14 \\ \hline
	  SEC \cite{SEC}& 0.39 & 0.16 & 0.27 & 0.13 & 0.35 & 0.19 & 0.07 & 0.22\\ \hline\hline
	 \multicolumn{9}{l}{keypoint supervision - category split} \\ \hline\hline
	 WTP \cite{WTP}& 0.16 & 0.14 & 0.20 & 0.20 & 0.01 & 0.07 & 0.13 & 0.13\\ \hline
	 \cite{cvpr} (VGG) & 0.46 & 0.48 & 0.72 & 0.78 & 0.44 & 0.53 & 0.65 & 0.58\\ \hline
	 Proposed (VGG)  & 0.55 & 0.48 & 0.72 & 0.76 & 0.49 & 0.48 & 0.67 & 0.59\\ \hline
	 \cite{cvpr} (ResNet) & 0.42 & 0.35 & 0.67 & 0.70 & 0.44 & 0.44 & 0.77 & 0.54\\ \hline
	 Proposed (ResNet) & 0.57 & 0.54 & 0.71 & 0.70 & 0.43 & 0.54 & 0.69 & 0.60\\ \hline\hline
	 \multicolumn{9}{l}{image label supervision - novel split} \\ \hline\hline
	  Area constraints \cite{weak_deeplab} & 0.05 & 0.00 & 0.04 & 0.16 & 0.00 & 0.01 & 0.32 & 0.09 \\ \hline
	 SEC \cite{SEC}& 0.12 & 0.03 & 0.06 & 0.23 & 0.07 & 0.12 & 0.25 & 0.13\\ \hline\hline
	 \multicolumn{9}{l}{keypoint supervision - novel split} \\ \hline\hline
	 WTP \cite{WTP}& 0.11 & 0.03 & 0.18 & 0.11 & 0.00 & 0.02 & 0.23 & 0.10\\ \hline
	 \cite{cvpr} (VGG) & 0.27 & 0.14 & 0.55 & 0.58 & 0.02 & 0.37 & 0.67 & 0.37\\ \hline
	  Proposed (VGG) & 0.31 & 0.18 & 0.56 & 0.49 & 0.08 & 0.41 & 0.66 & 0.38\\ \hline
	  \cite{cvpr} (ResNet) & 0.25 & 0.21 & 0.62 & 0.50 & 0.08 & 0.43 & 0.67 & 0.40\\ \hline
	  Proposed (ResNet) & 0.34 & 0.34 & 0.58 & 0.40 & 0.07 & 0.42 & 0.77 & 0.42\\ \hline
	 \end{tabular}
	 \caption{\label{tab:umd_dataset_results} Comparison of our method to the state-of-the-art on the UMD part affordance dataset. The Jaccard index is reported.}
	\end{table*}


\section{Conclusion}
In this work, we have proposed an approach for affordance segmentation that requires only weak supervision in the form of sparse keypoints. Our approach builds on the method \cite{cvpr}, but it does not require an additional graph cut segmentation step. This has been achieved by an adaptive approach for binarizing the predictions of a convolutional neural network during training. By approximating the Jaccard index based on the keypoints, we are also able to optimize parameters for the initialization. This approach could also be used to optimize other hyperparameters. We evaluated our approach on the CAD 120 affordance and the UMD part affordance dataset. Our approach outperforms the state-of-the-art for weakly supervised affordance segmentation. On the CAD 120 affordance dataset, the mean accuracy is increased by up to 17 percentage points compared to \cite{cvpr}.

\clearpage

{\small
\bibliographystyle{ieee}
\bibliography{egbib}
}

\end{document}